\def\year{2020}\relax
\documentclass[letterpaper]{article} 
\usepackage{aaai20}  
\usepackage{times}  
\usepackage{helvet} 
\usepackage{courier}  
\usepackage[hyphens]{url}  
\usepackage{graphicx} 

\usepackage{booktabs}
\usepackage{amsfonts,amssymb}
\usepackage{subcaption}
\usepackage{algorithm}
\usepackage{algorithmic}

\def\ie{\emph{i.e.~}}
\def\eg{\emph{e.g.~}}

\usepackage{graphicx}  
\title{A New Ensemble Adversarial Attack Powered by Long-term Gradient Memories}

\author{Zhaohui Che\footnotemark[2], Ali Borji\footnotemark[3], Guangtao Zhai\footnotemark[2]\thanks{Corresponding Author.}, Suiyi Ling\footnotemark[4], Jing Li\footnotemark[5], Patrick Le Callet\footnotemark[4]\\
\footnotemark[2]Shanghai Jiao Tong University, Shanghai, China\\
\footnotemark[3]MarkableAI Inc., Brooklyn, NY 11201 USA\\
\footnotemark[4]Universit{\'e} de Nantes, Nantes, France\\
\footnotemark[5]Alibaba Group, Hangzhou, China\\
 }

\begin{document}

\maketitle

\begin{abstract}
Deep neural networks are vulnerable to adversarial attacks. More importantly, some adversarial examples crafted against an ensemble of pre-trained source models can transfer to other new target models, thus pose a security threat to \textit{black-box} applications (when the attackers have no access to the target models). Despite adopting diverse architectures and parameters, source and target models often share similar decision boundaries. Therefore, if an adversary is capable of fooling several source models concurrently, it can potentially capture intrinsic transferable adversarial information that may allow it to fool a broad class of other \textit{black-box} target models.
Current ensemble attacks, however, only consider a limited number of source models to craft an adversary, and obtain poor transferability.
In this paper, we propose a novel \textit{black-box} attack, dubbed \textit{Serial-Mini-Batch-Ensemble-Attack} (\textit{SMBEA}). \textit{SMBEA} divides a large number of pre-trained source models into several mini-batches. For each single batch, we design 3 new ensemble strategies to improve the intra-batch transferability. Besides, we propose a new algorithm that recursively accumulates the ``long-term'' gradient memories of the previous batch to the following batch. This way, the learned adversarial information can be preserved and the inter-batch transferability can be improved. Experiments indicate that our method outperforms state-of-the-art ensemble attacks over multiple pixel-to-pixel vision tasks including image translation and salient region prediction. Our method successfully fools two online \textit{black-box} saliency prediction systems including DeepGaze-II \cite{deepgaze2online} and SALICON \cite{salicononline}. Finally, we also contribute a new repository to promote the research on adversarial attack and defense over pixel-to-pixel tasks: \url{https://github.com/CZHQuality/AAA-Pix2pix}.
\end{abstract}

\section{Introduction}
\label{sec:intro}

\begin{figure}
\centering
\includegraphics[height=0.45\linewidth]{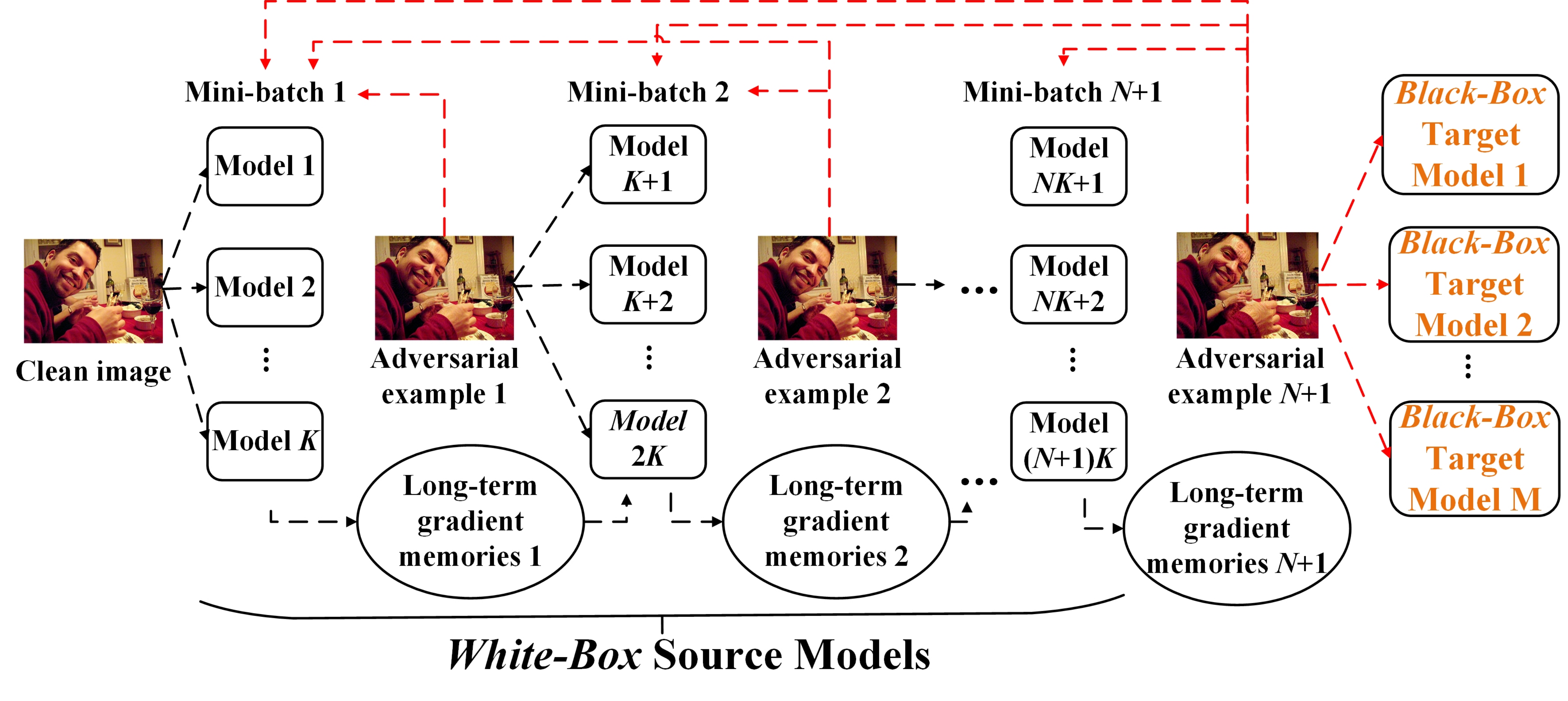}
\caption{\footnotesize The general idea of the proposed attack. Our attack divides a large number of pre-trained source models into several mini-batches. For each batch, we craft an adversary that fools multiple intra-batch source models. We also recursively accumulate the ``long-term'' gradient memories of previous batch to the following batch, in order to preserve the learned adversarial information and to improve inter-batch transferability. The red dashed lines denote that the crafted adversarial example can fool the previous source models, and also successfully fools the \textit{black-box} target models.}
\label{flowchart}
\vspace{-0.3cm}
\end{figure}

Deep neural networks, despite their great success in various vision tasks, are susceptible to adversarial attacks \cite{bfgs,fgsm}. The adversarial attacks add some quasi-imperceptible perturbations to the original input,
to significantly change the model output. More importantly, some well-designed adversarial examples can transfer across different models.
That is, the adversary crafted against some pre-trained source models can transfer to other new target models.
Despite the source and target models adopting diverse architectures and parameters, they may share similar decision boundaries. Thus, if an adversary can fool several source models, it can capture the intrinsic transferable adversarial information that allows it to fool a broad class of other \textit{black-box} target models.
The transferability of adversarial examples provides a potential chance to launch \textit{black-box} attacks without having access to the target model. In contrary, \textit{white-box} attacks require all information of target model, thus they are not practical in real world.

Particularly, adversarial attack serves as an efficient surrogate to evaluate the robustness of deep networks before they are deployed in real world, especially for security-related applications, \eg autonomous driving \cite{smautodrive,dataattmotodrive} and face verification \cite{AdvGlass,DongFaceAdv}. Therefore, exploring adversarial attacks, especially the transferable \textit{black-box} ones, is critical to demystifying the fragility of deep neural networks.

Current approaches for crafting transferable adversarial examples fall into two major categories: (1) \textit{Ensemble attacks} \cite{MIFGSM,EnsemAttack,iccvsegadv} craft transferable adversarial examples via fooling multiple \textit{white-box} source models in parallel.
(2) \textit{Generative methods} \cite{LatentGANAdv,GANimg} rely on an extra generative adversarial network (GAN). Specifically, a generator is trained to produce the adversaries that aim to fool the target model, while a discriminator is trained to distinguish the synthetic adversaries from original clean images for minimizing the perceptibility.

However, current methods have some drawbacks: (1) Normal \textit{ensemble attacks} only consider a limited number of source models.
To the best of our knowledge, the state-of-the-art \textit{ensemble attacks} \cite{EnsemAttack,MIFGSM} adopt less than 8 source models to craft the adversaries. In their implementations, all of the source models are combined in parallel. As a result, the number of source models is limited by the GPU memory.
(2) Although the parallel computing technique enables concurrent attacks against a large number of models, it brings new optimization challenges,
because computing and back-propagating the gradients of cost function w.r.t a large number of models become slow and difficult.
(3) \textit{Generative methods} rely on an extra GAN network which is not easy to train, and also require a lot of training samples with expensive labels.

For solving these problems, we propose a novel \textit{Serial-Mini-Batch-Ensemble-Attack} (\textit{SMBEA}). Before elaborating our method, we first introduce two empirical observations that inspire our method: (1) \textit{Crafting an adversarial example is analogous to training a model, and the transferability of the adversarial example is  analogous to the generalizability of the model} \cite{MIFGSM}. Thus, it is expected to increase the transferability of adversary via fooling diverse source models as much as possible, (2) \textit{Compared to the magnitude of the perturbation, the spatial structure of the adversarial perturbation has stronger impact on the final fooling ability} \cite{iccvsegadv}. Thus, we focus on preserving the learned adversarial structure information to optimize the fooling ability and transferability, while mitigating the magnitude of perturbation to minimize the perceptibility.

Inspired by the aforementioned empirical observations, our method mimics classical deep network training procedures to craft transferable adversaries, as shown in Fig.~\ref{flowchart}. Specifically, we divide a large number of pre-trained source models into several mini-batches, and each single batch contains $K$ ($K$ is the batch-size) individual source models. For each batch, we introduce 3 new ensemble strategies to combine these individual models, in order to improve intra-batch transferability.
For the inter-batch case, we propose a new algorithm that recursively accumulates the ``long-term'' gradient memories of previous batch to the following batch. This way, the learned adversarial information can be preserved and the inter-batch transferability can be improved.  As shown in Fig.~\ref{flowchart}, we start from a clean image, then recursively update the adversary across different batches, and finally obtain an adversary that not only fools all previous source models, but also fools new \textit{black-box} target models.

We summarize our contributions as follows:
\begin{itemize}
    \item {\bf A new black-box attack approach:} We propose a novel \textit{black-box} attack, where we introduce 3 new ensemble strategies for improving intra-batch transferability, and propose a new algorithm that preserves ``long-term'' gradient memories for improving inter-batch transferability.
    \item {\bf Generality:} Our method can attack multiple pixel-to-pixel vision tasks, \eg image translation and saliency prediction. Besides, our method successfully fools two online \textit{black-box} saliency prediction systems in the real world: \ie DeepGaze-II and SALICON. 
    \item {\bf A new repository:} We provide a \textit{software repository} including 13 common attack methods and our proposed attack, and 16 pre-trained source models. This repository aims to boost adversarial attack and defense research in pixel-to-pixel tasks. It also serves as a complement to \textit{CleverHans} repository \cite{CleverHans}.
\end{itemize}

\section{Related works}
In 2014, Szegedy \textit{et al.} verified the existence of adversarial examples for the first time \cite{bfgs}.

Goodfellow \textit{et al.} \cite{fgsm} introduced the fast gradient sign method (FGSM) to craft \textit{white-box} adversarial examples by one-step gradient update along the direction of the sign of gradient at each pixel.

Kurakin \textit{et al.} \cite{itefgsm} proposed the basic iterative version of FGSM, \ie I-FGSM. I-FGSM utilizes a small step to update adversarial example for multiple iterations by vanilla Stochastic Gradient Descent (\textit{SGD}) optimization. However, \textit{SGD} has some drawbacks, such as slow convergence and always drops into poor local minima.

Papernot \textit{et al.} \cite{PapernotBB} proposed a \textit{black-box} attack against image classifiers. Specifically, they trained a surrogate model to mimic the target \textit{black-box} model.

Dong \textit{et al.} \cite{MIFGSM} introduced the Momentum based Iterative Method (MIM), which utilizes Momentum based Stochastic Gradient Descent (\textit{MSGD}) \cite{MSGD} optimizer to craft adversaries. MIM accumulates the $1^{st}$ momentum in gradient descent direction to reduce poor local minima and to avoid ``over-fitting'' one specific model, thus demonstrating better transferability in \textit{black-box} setting.

Madry \textit{et al.} \cite{PGD} proposed the Projective Gradient Descent (PGD) attack, which extends the I-FGSM method to a universal first-order adversary by introducing a random start state. PGD also uses \textit{SGD} to update the adversary iteratively. It serves as a strong \textit{white-box} attack.

Carlini \textit{et al.} \cite{CandW} introduced an efficient \textit{white-box} attack, dubbed C$\&$W's attack, which breaks defensive distillation \cite{DefeseDis}. C$\&$W's attack utilizes \textit{Adam} optimization due to its fast convergence and high fooling ability.

Liu \textit{et al.} \cite{EnsemAttack} proposed \textit{targeted} and \textit{nontargeted} ensemble attacks that successfully fool \textit{black-box} classification system \ie \textit{Clarifai.com}. Wei \textit{et al.} \cite{GANimg} also trained a generative network to craft transferable adversaries against image and video detection models.

Xie \textit{et al.} \cite{iccvsegadv} proposed Dense Adversary Generation (DAG) to attack segmentation and object detection models.
Mopuri \textit{et al.} \cite{pamifeature} introduced a general objective function that produces image-agnostic adversaries from latent space.

\begin{figure*}
\centering
\includegraphics[height=0.20\linewidth]{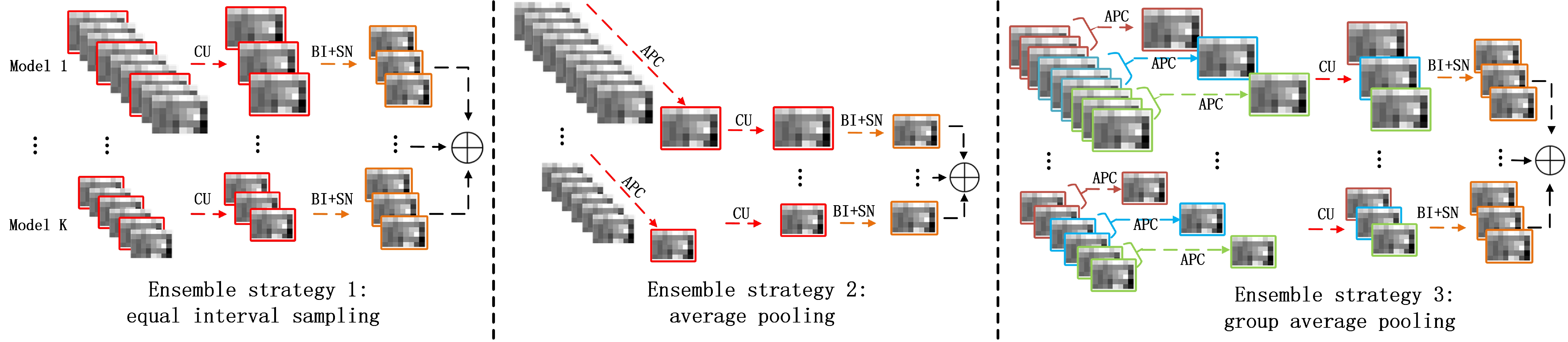}
\caption{\footnotesize Visualizations of three ensemble strategies in \textit{feature space}. The raw feature maps are processed by the batch normalization and ReLU activation function. \textbf{CU}: Channel amount Unification, \textbf{BI}: Bilinear Interpolation, \textbf{SN}: Softmax Normalization, \textbf{APC}: Average Pooling in Channel direction. $\oplus$ represents an elementwise weighted summation.}
\label{ensemblestr}
\end{figure*}

\section{Proposed method}
In this section, we introduce a novel \textit{black-box} adversarial attack dubbed \textit{Serial-Mini-Batch-Ensemble-Attack} (\textit{SMBEA}). We first introduce the intra-batch algorithm, then we elaborate the inter-batch algorithm.

\subsection{Intra-batch ensemble strategies}
In our implementation, each single mini-batch includes $K$=4 \footnote{\scriptsize Batch-size $K$ is a tunable hyper-parameter. Here we adopt 4 as the batch-size because it achieves a good tradeoff between the transferability and GPU memory cost.} \textit{white-box} source models that are pre-trained over the same pixel-to-pixel task. Thus, these models have similar decision boundaries, despite adopting diverse architectures and parameters.
Similar decision boundaries across models increase the chance of crafting an adversary that fools all these models. At the same time the diversity across different models serves as the regularization and alleviates ``over-fitting'' to a specific model, which in turn results in high intra-batch transferability.

In this work, we only consider the \textit{targeted attack}. The \textit{non-targeted attack} is a straightforward extension.
We formulate the \textit{targeted} ensemble attack in pixel-to-pixel tasks as a constrained optimization problem.
For simplicity, we first introduce a basic ensemble strategy, where multiple models are fused in \textit{output space}, \ie the optimization objective is computed by an element-wise weighted summation of the final predictions of multiple source models:
\begin{equation}
\footnotesize
\left\{
             \begin{array}{lr}
             \textup{min}~\mathcal L_o = \mathcal L_1 [\sum\limits_{n=1}^{K} \sigma_n \cdot \mathbb{F}_{n}({ I^*}), ~\mathcal F (G)] + \lambda_1 \cdot \mathcal L_2 ({ I}, ~~{ I^*}), \\
             s.t. ~~~~~~\mathcal L_2 ({ I}, ~~{ I^*}) \leq \mathcal T_1.
             \end{array}
\right.
\label{eqoutspc}
\end{equation}
where $I$ and $I^*$ are original clean image and adversarial example, respectively. $G$ represents the guide image, while $\mathcal F (G)$ is the ground-truth output of $G$. For the \textit{targeted attack}, the goal is to change the models' ensemble prediction of $I^*$ towards the prediction of the guide image.
$\mathbb{F}_n$ is the $n_{th}$ source model within the mini-batch, and $\mathbb{F}_{n}({ I^*})$ represents the final prediction of the $n_{th}$ model on the crafted adversary $I^*$.
$\sigma_n$ is ensemble weight, $\sum_{n=1}^K \sigma_n = 1$.
$\mathcal L_1$ is the loss function that is minimized when $\sum_{n=1}^{K} \sigma_n \cdot \mathbb{F}_{n}({ I^*}) = \mathcal F (G)$. $\mathcal L_2$ is the perceptual constraint, \textit{e.g.} {$\bf L_{0}$}, {$\bf L_{1}$}, or {$\bf L_{\infty}$} norms, which is minimized to guarantee that the crafted adversary ${I^*}$ looks (perceptually) similar to the original clean image $I$. $\mathcal T_1$ is the maximum perceptual constraint for single batch. $\lambda_1$ is a hyper-parameter to balance the fooling ability loss $\mathcal L_1$ and the perceptual constraint $\mathcal L_2$.

However, the basic ensemble strategy in \textit{Eq}.~\ref{eqoutspc} only fools the final predictions in the \textit{output space}. Here we further dig into the \textit{feature space} to explore other efficient ensemble strategies. This is motivated by the consideration that combining \textit{output space} and \textit{feature space} ensembles provides a deep supervision for
crafting strong adversary that not only fools the final predictions, but also fools the intermediate feature maps.
This way, the objective $\mathcal L_{o,f}$ is rewritten as:
\begin{equation}
\footnotesize
\left\{
             \begin{array}{lr}
             \textup{min}~\mathcal L_{o,f} = \mathcal L_{o} + \lambda_2 \cdot \mathcal L_3 [\sum\limits_{n=1}^{K} \omega_n \cdot \mathbb{D}_{n}({ I^*}), \sum\limits_{n=1}^{K} \omega_n \cdot \mathbb{D}_{n}({ G})],\\
             s.t. ~~\mathcal L_2 ({ I}, ~~{ I^*}) \leq \mathcal T_1.
             \end{array}
\right.
\label{eqfeaspc}
\end{equation}
where $\mathbb{D}_n(I^*)$ and $\mathbb{D}_n(G)$ represent the feature maps of the $n_{th}$ source model on the crafted adversary $I^*$ and guide image $G$. $\omega_n$ is the \textit{feature space} ensemble weight, $\sum_{n=1}^K \omega_n = 1$. $\mathcal L_3$ is the loss function that aims to minimize the \textit{feature space} distance between $I^*$ and $G$. $\lambda_2$ is a hyper-parameter to balance \textit{feature space} fooling loss $\mathcal L_3$, together with \textit{output space} fooling loss and perceptual constraint.

Different models utilize different network architectures, so their feature maps have different resolutions and channels. For solving this, we introduce 3 different \textit{feature space} ensemble strategies, as shown in Fig.~\ref{ensemblestr}. More details regarding the feature layer selection are provided in the experiments section. Here we focus on explaining ensemble strategies.

The first ensemble strategy evenly samples $p$ feature maps from each model, and the sampling interval $\bar p_n$ of the $n_{th}$ model can be computed as $\bar p_n = P_n / p$, where $P_n$ is the total number of feature channels of one selected feature layer from the $n_{th}$ model. We set $p$ and $P_n$ as the powers of 2, to make sure that $\bar p_n$ is an integer. This way, we obtain the same number of feature maps (channels) from different models. Next, we use bilinear interpolation to resize the selected feature maps of different models to the same resolution (\ie height$\times$weight). Then, we adopt \textit{softmax} function to normalize these feature maps. Finally, we obtain the \textit{feature-space} ensemble result by an element-wise weighted summation of different feature maps.

The other two ensembles have similar pipelines, except for the first step, which is explained below.
For each model, the second ensemble computes the average pooling of the $P_n$ feature maps in the channel direction, and obtains a 2D one-channel feature map from each model.
The third ensemble divides $P_n$ feature maps into $p$ groups, then computes the channel direction average of each group, and obtains $p$ candidate feature maps from each model.

\begin{algorithm}[htb]
\scriptsize
\caption{: \scriptsize Intra-batch update rules of \textit{SMBEA}. This algorithm is applicable to the first mini-batch. $m_t$ represents the $1^{st}$ gradient momentum vector, while $v_t$ represents the $2^{nd}$ gradient momentum vector. $\odot$ is an element-wise product.}
\label{alg:intra}
\begin{algorithmic}[1]
\REQUIRE ~~\\
Original clean image $I$, ~~guide image $G$ (randomly selected);\\
Intra-batch source models: $\mathbb{F}_1$, $\mathbb{F}_2$, ..., $\mathbb{F}_K$; \\
Decay factors of short-term gradient momentums: $\mu_1$, $\mu_2$; \\
Smoothing term: $\epsilon$;\\
Maximum iterations \textit{X} for single mini-batch;\\
Maximum perceptual constraint $\mathcal T_1$ for the first mini-batch; \\
Step size of iterative gradient descent $\alpha$;\\
\ENSURE ~~\\
An adversarial example $I^*_X$; ~~ The ultimate $1^{st}$ momentum $m_X$, ~~ and the ultimate $2^{nd}$ momentum $v_X$.\\
\label{ code:fram:extract }
\STATE Initialization: $I^*_0 \leftarrow I$, $m_0 \leftarrow \textbf{0}^d$, $v_0 \leftarrow \textbf{0}^d$, $t \leftarrow 0$ \\
\label{code:fram:trainbase}
\WHILE{($0 \leq t < \textit{X}$ ~~and~~ $\|{ I}, ~~{ I^*_{t}}\|_1 \leq \mathcal T_1$)}
\STATE $t \leftarrow t+1$; ~~~(update the iteration epoch)\\
\STATE $g_t \leftarrow \nabla_{I^*_{t-1}} \mathcal L_{o,f}$; ~~~($\mathcal L_{o,f}$ is defined in \textit{Eq}.~\ref{eqfeaspc}) \\
\STATE $\hat{g_t} \leftarrow \frac{g_t}{\|g_t\|_1};$ ~~~(gradient normalization)\\
\STATE $m_t \leftarrow \mu_1 \cdot m_{t-1} + (1-\mu_1) \cdot \hat{g_t}$; ~~~(update the $m_t$)\\
\STATE $v_t \leftarrow \mu_2 \cdot v_{t-1} + (1-\mu_2) \cdot \hat{g_t}^2;$ ~~~(update the $v_t$)\\
\STATE $\hat{m_t} \leftarrow m_t/(1-\mu_1^t)$; ~~~(bias correction)\\
\STATE $\hat{v_t} \leftarrow v_t/(1-\mu_2^t)$; ~~~(bias correction)\\
\STATE $I^*_t \leftarrow \textup{Clip}(I^*_{t-1} - \alpha \cdot \frac{1}{\sqrt{\hat{v_{t}}} + \epsilon} \odot \hat{m_t})$; ~~~(update the adversary)\\
\ENDWHILE
\RETURN $I^*_X \leftarrow I^*_t$, ~~$m_X \leftarrow m_t$, ~~$v_X \leftarrow v_t$.
\end{algorithmic}
\end{algorithm}

\subsection{Intra-batch update rules}

For solving the constrained optimization problem in \textit{Eq}.~\ref{eqfeaspc}, we exhaustively test 5 common gradient descent optimization methods, \ie stochastic gradient descent (\textit{SGD}), momentum based gradient descent (\textit{MSGD}) \cite{MSGD}, \textit{Adagrad} \cite{Adagrad}, \textit{RMSProp} \cite{RMSProp}, and \textit{Adam} \cite{Adam}. The major differences between these gradient descent methods are two gradient momentums, explained below.

The $1^{st}$ gradient momentum accumulates the gradients of previous iterations to stabilize the gradient descent direction, and helps to reduce poor local minima.

The $2^{nd}$ gradient momentum adapts the learning rates to different parameters. In pixel-to-pixel attacking tasks, we aim to update the image pixels of the crafted adversarial example.
However, in the attacking process, a small fraction of pixels are frequently updated, while the remaining pixels are occasionally updated. The intensities of frequent pixels grow rapidly and reach the bound of perceptual constraint quickly, while the infrequent pixels are far from convergence at this moment. This issue limits the tradeoff between fooling ability and perceptibility. For mitigating this limitation, the $2^{nd}$ gradient momentum was proposed, which assigns a small update step-size for frequent pixels, while assigning a big update step-size for infrequent pixels.

In our tasks, \textit{Adam} achieves the best tradeoff between transferability, perceptibility, and convergence speed.
The update rules of the intra-batch algorithm based on \textit{Adam} optimizer are given in the Algorithm ~\ref{alg:intra}.
Specifically, \textit{Adam} optimizer utilizes the $1^{st}$ momentum to avoid local poor minima and to prevent ``over-fitting'', thus improving transferability. Besides, it also uses the $2^{nd}$ momentum to adapt the learning rates for different pixels, thus enhancing the tradeoff between fooling ability and perceptibility.

In our implementation, the original image $I$ and the guide image $G$ are normalized to be in the range [0, 1]. The default decay factors are set as $\mu_1 = 0.9$ and $\mu_2 = 0.99$. $\epsilon = 1 \times 10^{-8}$ is a smoothing term to avoid division by zero. The maximum number of iterations \textit{X} for a single batch is 20. The iterative gradient descent step size $\alpha=2\times10^{-4}$.
We adopt $\bf L_1$ norm as the perceptual constraint.
Finally, we clip the crafted adversary $I^*_t$ into the range [0, 1] to make sure $I^*_t$ is a valid image.
This way, we obtain an adversary $I^*$ that is able to fool multiple intra-batch source models.

\subsection{Inter-batch update rules}
The intra-batch algorithm only guarantees that the crafted adversary can fool a limited number of source models. For breaking this limitation, we propose a novel inter-batch algorithm that recursively accumulates the ``long-term'' gradient memories of the previous batches to the following batches. These ``long-term'' gradient memories preserve the learned adversarial information, and also serve as the regularization to prevent ``over-fitting'' on a specific batch, thus increasing the inter-batch transferability.

The proposed inter-batch update rules are presented in the Algorithm~\ref{alg:inter}. The main differences between the two algorithms are the initialization and the bias corrections.

\begin{algorithm}[htb]
\scriptsize
\caption{: \scriptsize Inter-batch update rules of \textit{SMBEA}. This algorithm is applicable to all mini-batches, except for the first one, \ie $i>1$. Notice that the superscript in brackets denotes the batch number, \eg $I^{*~(i)}_X$ is the adversary of the $i_{th}$ batch, while the superscript w/o brackets denotes the pow, \eg  $\beta_3^i$ denotes the $\beta_3$ to the power $i$.}
\label{alg:inter}
\begin{algorithmic}[1]
\REQUIRE ~~\\
The adversarial example of the previous batch $I^{*~(i-1)}_X$, ~~the guide image $G$;\\
The $1^{st}$ gradient momentum of the previous batch $m_X^{(i-1)}$;\\
The $2^{nd}$ gradient momentum of the previous batch $v_X^{(i-1)}$;\\
Maximum perceptual constraint $\mathcal T_1^{(i-1)}$ of the previous batch;\\
Maximum perceptual constraint $\mathcal T_1^{(1)}$ of the first batch;\\
Maximum batch number $N$;\\
Intra-batch models of the current batch: $\mathbb{F}_1^{(i)}$, $\mathbb{F}_2^{(i)}$, ..., $\mathbb{F}_K^{(i)}$; \\
Decay factors of short-term gradient momentums: $\mu_1$, $\mu_2$; \\
Weights of long-term gradient momentums: $\beta_1$, $\beta_2$ $\in [0, 1]$; \\
Decay factor of perceptual constraint: $\beta_3 \in [0, 1]$; \\
\ENSURE ~~\\
An adversarial example of current batch $I^{*~(i)}_X$;~~ The ultimate $1^{st}$ momentum $m^{(i)}_X$, ~~and ultimate $2^{nd}$ momentum $v^{(i)}_X$ of the current batch.\\
\label{ code:fram:extract }
\STATE Initialization: $I^{*~(i)}_0 \leftarrow I^{*~(i-1)}_X$, ~~~$m_0^{(i)} \leftarrow \beta_1 \cdot m_X^{(i-1)}$, \\~~~$v_0^{(i)} \leftarrow \beta_2 \cdot v_X^{(i-1)}$, ~~~$\mathcal T_1^{(i)} \leftarrow \mathcal T_1^{(i-1)} + \beta_3^i \cdot \mathcal T_1^{(1)}$, ~~~$t \leftarrow 0$ \\
\label{code:fram:trainbase}
\WHILE{($i \leq N$ ~~and~~ $0 \leq t < \textit{X}$ ~~and~~ $\|{ I}, ~~{ I^*_{t}}\|_1 \leq \mathcal T_1^{(i)}$)}
\STATE do Step.3~-~Step.7 of the Algorithm. 1. \\
\STATE $\hat{m_t}^{(i)} \leftarrow \frac{m_t^{(i)}}{(1-\mu_1^t)} + \beta_1 \cdot m_X^{(i-1)}$; ~~~(bias correction)\\
\STATE $\hat{v_t}^{(i)} \leftarrow \frac{v_t^{(i)}}{(1-\mu_2^t)} + \beta_2 \cdot v_X^{(i-1)}$; ~~~(bias correction)\\
\STATE $I^{*~(i)}_t \leftarrow \textup{Clip}(I^{*~(i)}_{t-1} - \alpha \cdot \frac{1}{\sqrt{\hat{v}^{(i)}_{t}} + \epsilon} \odot \hat{m}^{(i)}_t)$;\\
\ENDWHILE
\RETURN $I^{*~(i)}_X \leftarrow I^{*~(i)}_t$, $m^{(i)}_X \leftarrow m^{(i)}_t$, $v^{(i)}_X \leftarrow v^{(i)}_t$, $i \leftarrow i+1$.
\end{algorithmic}
\end{algorithm}

We first introduce the initialization. We adopt four variables of the previous batch to recursively initialize the variables of the following batch, explained below.
\begin{itemize}
\item \textit{$I^{*~(i)}_0 \leftarrow I^{*~(i-1)}_X$}: we adopt the adversarial example of the previous batch $I^{*~(i-1)}_X$ as the initial state of the current batch, because $I^{*~(i-1)}_X$ has learned some adversarial information against multiple models of previous batch.
\item \textit{$m_0^{(i)} \leftarrow \beta_1 \cdot m_X^{(i-1)}$ and $v_0^{(i)} \leftarrow \beta_2 \cdot v_X^{(i-1)}$}: we utilize the $1^{st}$ and $2^{nd}$ momentums of the previous batch to initialize the momentums of the current batch. These ``long-term'' gradient momentums preserve the learned adversarial information, and also serve as the regularization to prevent ``over-fitting'' on the following batch, thus boosting the inter-batch generalizability of the crafted adversary. %
\item \textit{$\mathcal T_1^{(i)} \leftarrow \mathcal T_1^{(i-1)} + \beta_3^i \cdot \mathcal T_1^{(1)}$}: we recursively update the maximum perceptual constraint of the current batch (\ie $\mathcal T_1^{(i)}$) by adding a loose factor $\beta_3^i \cdot \mathcal T_1^{(1)}$ to the perceptual constraint of the previous batch (\ie $\mathcal T_1^{(i-1)}$), in order to prevent premature convergence that causes ``under-fitting''. Besides, by increasing the number of batches, the adversarial example tends to be converged, so we reduce the loose factor via a decay rate $\beta_3^i$, where $\beta_3^i$ denotes the $\beta_3 \in [0, 1]$ to the power $i$ ($i$ is the batch number).
\end{itemize}

The proposed inter-batch algorithm inherits the good properties of the classical \textit{Adam} method, explained below.

\textbf{Property 1:}
\textit{The effective step-size of inter-batch update rules is invariant to the scale transform of the gradient.}

\textbf{Proof 1:}
As shown in Step.6 of Algorithm.~\ref{alg:inter}, assuming $\epsilon=0$, the effective step-size of the adversarial example at iteration $t$ is $\Delta_t^{(i)} = \alpha \cdot \frac{1}{\sqrt{\hat{v}^{(i)}_{t}}} \odot \hat{m}^{(i)}_t$.
The effective step-size $\Delta_t^{(i)}$ is invariant to the scale transform of gradient, because scaling raw gradient $g_t$ with factor $c$ will be normalized by $\bf L_1$ norm, \ie $\frac{g_t}{\|g_t\|_1} = \frac{c \cdot g_t}{\|c \cdot g_t\|_1}$. Thus, $\hat{m_t}^{(i)}$, $\hat{v_t}^{(i)}$, $\Delta_t^{(i)}$ are invariant to the scale transform of the gradient.

\textbf{Property 2:}
\textit{The proposed inter-batch bias corrections can correct for the discrepancy between the expected value of the exponential moving averages (\ie $\mathbb E[m_t^{(i)}]$ or $\mathbb E[v_t^{(i)}]$) and the true expected gradients (\ie $\mathbb E[\hat{g_t}]$ or $\mathbb E[\hat{g_t}^2]$)}.

\textbf{Proof 2:}
The proposed inter-batch bias corrections are shown in Steps.4-5 of the Algorithm.~\ref{alg:inter}. Here, we derive the bias correction for the $2^{nd}$ momentum estimate, and the derivation for the $1^{st}$ momentum is completely analogous.

Let $\hat{g_t}$ be the normalized gradient at iteration $t$, and we wish to estimate its $2^{nd}$ momentum $\hat{v}_t^{(i)}$ using an exponential moving average of the true squared gradient. In the inter-batch case, the raw $2^{nd}$ momentum is initialized as $v_0^{(i)} \leftarrow \beta_2 \cdot v_X^{(i-1)}$. The recursive update equation of raw momentum $v_t^{(i)} = \mu_2 \cdot v_{t-1}^{(i)} + (1-\mu_2) \cdot \hat{g_t}^2$ can be rewritten as:
\begin{equation}
\footnotesize
v_t^{(i)} = \mu_2^t \cdot v_0^{(i)} + (1-\mu_2)\sum_{k=1}^{t}\mu_2^{t-k} \cdot \hat{g_k}^{2},
\label{secupdate}
\end{equation}

We wish to know how $\mathbb E[v_t^{(i)}]$, the expected value of the exponential moving average at iteration $t$, relates to the true expected squared gradient $\mathbb E[\hat{g_t}^2]$, so we can correct for the discrepancy between them. We take expectations of the left and right sides of \textit{Eq}.~\ref{secupdate}
\begin{equation}
\small
\left\{
             \begin{array}{lr}
             \mathbb E [v_t^{(i)}] = \mathbb E [\mu_2^t \cdot v_0^{(i)}] + \mathbb E [(1-\mu_2)\sum_{k=1}^{t}\mu_2^{t-k} \cdot \hat{g_k}^2] \\
             ~~~~~~~~~~~= \mu_2^t \cdot v_0^{(i)} + \mathbb E[\hat{g_t}^2] \cdot (1-\mu_2) \sum_{k=1}^{t} \mu_2^{t-k} + \zeta \\
             ~~~~~~~~~~~= \mu_2^t \cdot v_0^{(i)} + \mathbb E[\hat{g_t}^2] \cdot (1-\mu_2^t) + \zeta,
             \end{array}
\right.
\label{secupdate2}
\end{equation}
where $\zeta=0$ if the true $2^{nd}$ momentum $ \mathbb E[\hat{g_t}^2]$ is stationary, according to \textit{Adam} \cite{Adam}.
We suppose $\mathbb E[\hat{g_t}^2]$ is stationary, and divide the left and right sides of \textit{Eq}.~\ref{secupdate2} by $(1-\mu_2^t)$:
\begin{equation}
\footnotesize
\frac{\mathbb E[v_t^{(i)}]}{1-\mu_2^t} = \frac{\mu_2^t \cdot v_0^{(i)}}{1-\mu_2^t} + \mathbb E[\hat{g_t}^2],
\label{secupdate3}
\end{equation}
where $\frac{\mu_2^t \cdot v_0^{(i)}}{1-\mu_2^t} = \frac{\mu_2^t}{1-\mu_2^t} \cdot \beta_2 \cdot v_X^{(i-1)}$ ($v_0^{(i)}$ is initialized as $\beta_2 \cdot v_X^{(i-1)}$) is the ``long-term'' momentum from the previous batch. This ``long-term'' momentum decreases rapidly with the increase of iteration $t$ due to the decay rate $\frac{\mu_2^t}{1-\mu_2^t}$. However, in our tasks, we wish to assign a smooth decay rate to the ``long-term'' momentum in subsequent iterations, in order to preserve the learned adversarial information as much as possible.
Thus, we modify the decay weight as $\frac{1}{1-\mu_2^t}$, which obtains a slower decay rate than $\frac{\mu_2^t}{1-\mu_2^t}$. We first subtract $ \frac{\mu_2^t \cdot v_0^{(i)}}{1-\mu_2^t}$, then add $\frac{1 \cdot v_0^{(i)}}{1-\mu_2^t}$ to the left and right sides of \textit{Eq.}~\ref{secupdate3}
\begin{equation}
\footnotesize
\frac{\mathbb E[v_t^{(i)}]}{(1-\mu_2^t)} - \frac{\mu_2^t \cdot v_0^{(i)}}{1-\mu_2^t} + \frac{1 \cdot v_0^{(i)}}{1-\mu_2^t} = \frac{1 \cdot v_0^{(i)}}{1-\mu_2^t} + \mathbb E[\hat{g_t}^2],
\label{secupdate4}
\end{equation}
Next, we plug $v_0^{(i)} = \beta_2 \cdot v_X^{(i-1)}$ into \textit{Eq}.~\ref{secupdate4}, and obtain
\begin{equation}
\footnotesize
\frac{\mathbb E[v_t^{(i)}]}{(1-\mu_2^t)} + \beta_2 \cdot v_X^{(i-1)} = \frac{1 }{1-\mu_2^t} \cdot \beta_2 \cdot v_X^{(i-1)} + \mathbb E[\hat{g_t}^2].
\label{secupdate5}
\end{equation}
This way, we obtain the corrected $2^{nd}$ momentum, as shown in Step.5 of Algorithm.~\ref{alg:inter} (left side of \textit{Eq}.\ref{secupdate5}), which is composed of two parts (right side of \textit{Eq}.\ref{secupdate5}): the ``long-term'' gradient momentum with a smooth decay rate $\frac{1 }{1-\mu_2^t} \cdot \beta_2 \cdot v_X^{(i-1)}$, and the true expected squared gradient $\mathbb E[\hat{g_t}^2]$.

We utilize 3 new hyper-parameters in the inter-batch algorithm, \ie $\beta_1$, $\beta_2$, and $\beta_3$, where $\beta_1$ and $\beta_2$ control the weights of ``long-term'' momentums from previous batches, and $\beta_3$ decides the decay rate of the perceptual constraint. In our implementation, the default settings are $\beta_1=0.10$, $\beta_2=0.01$, $\beta_3=0.60$. For selecting the good settings, we test these hyper-parameters by line-searching on 2 validation datasets, \ie Cityspaces and LSUN'17.

\begin{table*}
\centering
\small
    \caption{\footnotesize Comparison under \textit{black-box} setting. Fooling ability is measured by performance drop: \textit{CC}$\uparrow$ for LSUN'17, \textit{MSE}$\downarrow$ for other datasets.}
    \label{Comparison}
    \scriptsize
    \centering
    \begin{tabular}{l|c|c|c|c}
\toprule[1pt]
    {Datasets (original performance)} &{ Cityspaces ~(\textit{MSE}=0.0139)} &  Facades ~(\textit{MSE}=0.0521) &  Google Satellite ~(\textit{MSE}=0.0255) &  LSUN'17 ~(\textit{CC}=0.7748)  \\
   \hline
   {Target \textit{black-box} model} &{ pix2pix U-Net} &  Global pix2pixHD & Local pix2pixHD & SALICON \\
   \hline
   {Number of mini-batches}    &{ 1 ~~~~~~~~~~~~~ 3 ~~~~~~~~~~~~~ 5} &{ 1 ~~~~~~~~~~~~~ 3 ~~~~~~~~~~~~~ 5} &{ 1 ~~~~~~~~~~~~~ 3 ~~~~~~~~~~~~~ 5} &{ 1 ~~~~~~~~~~~~~ 3 ~~~~~~~~~~~~~ 5}\\
   \hline
   {Percep. cons. ($\bf L_1$ norm)}    &{ 1.0$e^{-2}$ ~~~~ 2.0$e^{-2}$ ~~~~ 2.4$e^{-2}$} &{ 1.2$e^{-2}$ ~~~~ 2.4$e^{-2}$ ~~~~ 2.8$e^{-2}$} &{ 1.0$e^{-2}$ ~~~~ 2.0$e^{-2}$ ~~~~ 2.4$e^{-2}$} &{ 3.4$e^{-2}$ ~~~~ 6.6$e^{-2}$ ~~~~ 7.8$e^{-2}$}\\
   \hline
   \hline
{Random noise} &{ +0.0002 ~~~~ +0.0008 ~~~~ +0.0011}   &{ +0.0003 ~~~~ +0.0010 ~~~~ +0.0013} &{ +0.0002 ~~~~ +0.0006 ~~~~ +0.0011} &{ -0.0002 ~~~~ -0.0002 ~~~~ -0.0003} \\
{Ensemble Attack using PGD } &{ +0.0108 ~~~~ +0.0169 ~~~~ +0.0174}   &{ +0.0104 ~~~~ +0.0133 ~~~~ +0.0166} &{ +0.0074 ~~~~ +0.0083 ~~~~ +0.0099} &{ -0.0022 ~~~~ -0.0264 ~~~~ -0.0511} \\
{Ensemble Attack using C$\&$W } &{ +0.0113 ~~~~ +0.0170 ~~~~ +0.0173 }   &{ +0.0097 ~~~~ +0.0131 ~~~~ +0.0168} &{ +0.0068 ~~~~ +0.0087 ~~~~ +0.0096} &{ -0.0763 ~~~~ -0.2117 ~~~~ -0.2452} \\
{Ensemble Attack using MIM } &{ +0.0116 ~~~~ +0.0193 ~~~~ +0.0227}   &{ +0.0125 ~~~~ +0.0162 ~~~~ +0.0178} &{ +0.0093 ~~~~ +0.0099 ~~~~ +0.0113} &{ -0.0771 ~~~~ -0.2417 ~~~~ -0.2880} \\
{Liu's Ensemble Attack} &{ +0.0118 ~~~~ +0.0194 ~~~~ +0.0230}   &{ +0.0129 ~~~~ +0.0165 ~~~~ +0.0184} &{ +0.0098 ~~~~ +0.0105 ~~~~ +0.0116} &{ -0.0780 ~~~~ -0.2533 ~~~~ -0.2941} \\
{Proposed \textit{SMBEA}} &{\bfseries +0.0148 ~~~~ +0.0213 ~~~~ +0.0264}   &{\bfseries +0.0155 ~~~~ +0.0230 ~~~~ +0.0275 } &{\bfseries +0.0108 ~~~~ +0.0137 ~~~~ +0.0145} &{\bfseries -0.0871 ~~~~ -0.3017 ~~~~ -0.4180} \\
\toprule[1pt]
    \end{tabular}
\end{table*}

\begin{figure*}
\centering
\includegraphics[height=0.3\linewidth]{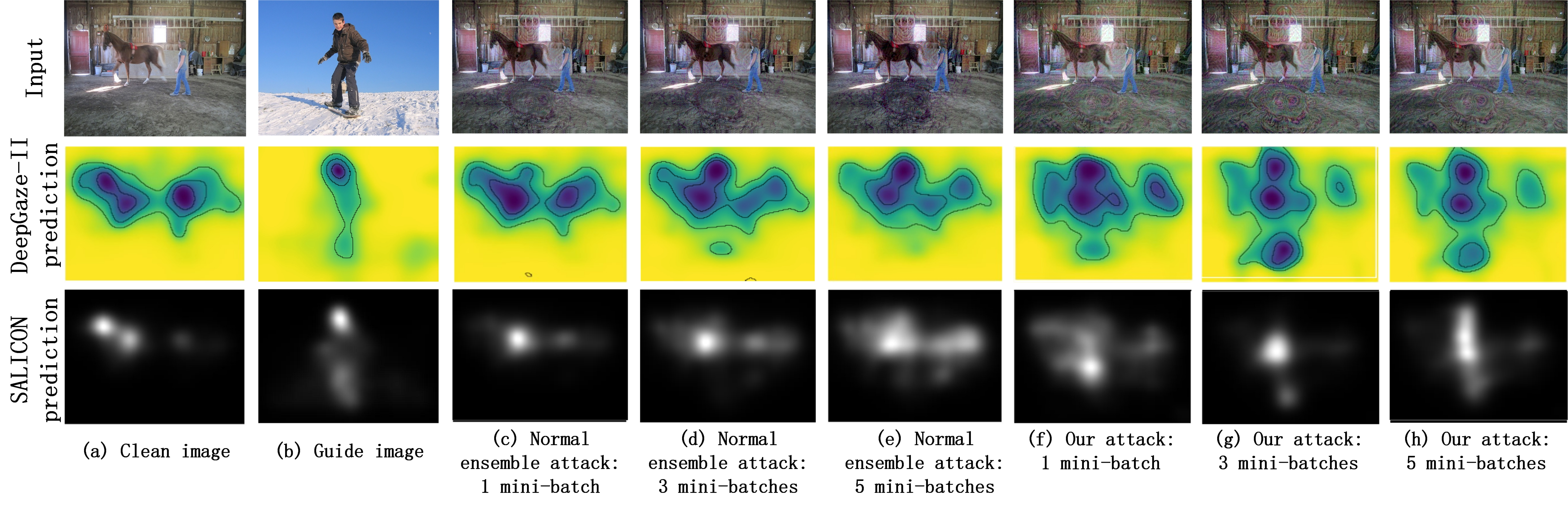}
\caption{\footnotesize Attacking real-world applications. With the increase of batch number, our attack fools two online \textit{black-box} saliency prediction systems, \ie DeepGaze-II and SALICON. However, normal ensemble attack based on PGD \cite{PGD} fails to fool these models.}
\label{olattack}
\end{figure*}

\section{Experiments}
\subsection{The selection of loss functions}
In \textit{Eqs}.\ref{eqoutspc}-\ref{eqfeaspc}, we provide a general paradigm for computing objective cost functions. For different tasks, we select different task-specific loss metrics to reach a better attack performance. Specifically, for image-translation, we use a linear combination of \textit{Mean Absolute Error} (\textit{MAE}), negative \textit{Pearson's Linear Correlation Coefficient} (\textit{CC}), and VGG loss \cite{PercepLoss} as the \textit{output space} fooling ability loss $\mathcal L_1$. For saliency prediction, we use a linear combination of \textit{Kullback-Leibler divergence} (\textit{KL}), \textit{MAE}, and negative \textit{CC} as $\mathcal L_1$. We use \textit{KL} as \textit{feature-space} fooling ability loss $\mathcal L_3$, because the intermediate feature maps of source models are normalized by \textit{softmax} as the distributions. The averaged $\bf L_1$ norm serves as perceptual loss $\mathcal L_2$.

\subsection{Datasets and evaluation protocol}
To explore the generalization ability of \textit{SMBEA},
we conduct experiments on 4 pixel-to-pixel vision datasets, \ie Cityspaces \cite{Cityspaces}, Facades \cite{Facades}, Google Satellites, and LSUN'17 \cite{SALICONDB}.
For Cityspaces, we select 1000 ``\textit{Semantic label} $\&$ \textit{Real photo}'' pairs as test set. For Facades, we select 400 ``\textit{Architectural label} $\&$ \textit{Real photo}'' pairs. For Google Satellites, we select 1000 ``\textit{Google Map} $\&$ \textit{Satellite Image}'' pairs. For LSUN'17, we select 1000 ``\textit{Real Photo} $\&$ \textit{Saliency Map}'' pairs.

For fair comparison, we adopt the performance drop to measure the fooling ability, \ie the difference between the performances on clean images and on adversarial examples. The stronger the attack, the bigger the performance drop. For image translation tasks, we use the \textit{Mean Squared Error} (\textit{MSE}) to measure the performance drop.
For the saliency prediction task, we use the \textit{Pearson's Linear Correlation Coefficient} (\textit{CC}) to measure the performance drop.
For measuring the perceptibility, we use the averaged $\bf L_1$ norm. In this paper, the images are normalized to be in the range [0, 1], and the $\bf L_1$ norm is averaged by the number of pixels.

\subsection{Source models}
For the saliency prediction task, we adopt 16 state-of-the-art deep saliency models as the raw source models. However, we wish to use additional source models to explore the upper-bound of our attack. Thus, we design an augmentation strategy to enlarge the number of models.
Specifically, we replace the standard convolution (adopted in most of the current CNN models) in the original architecture with two new convolutions, \ie deformable convolution kernel \cite{DCN1} and dilated convolution kernel \cite{DilatedConv}. By doing so, for each raw source model, we obtain two new variants that have diverse architectures, without causing obvious performance drop. This way, we obtain 48 source models in total. We use the same model augmentation strategy for other tasks.

\begin{table}
\centering
\small
\caption{\footnotesize Evading defense: attack performance comparison against the adversarially trained \textit{black-box} models. LSUN'17 is the test set.}
\label{AgaintDef}
    \scriptsize
    \centering
    \begin{tabular}{c|c|c|c|c}
\toprule[1pt]
{Target model} &{Attack}   &{No. of Batch: 3} &{No. of Batch: 5} &{No. of Batch: 7} \\
\hline
{GazeGAN} &{I-FGSM}   &{-0.0125} &{-0.0177} &{-0.0206} \\

{GazeGAN} &{MIM}   &{-0.1315} &{-0.1582} &{-0.2063} \\

{GazeGAN} &{\textit{SMBEA}}   &{\textbf{-0.2190}} &{\textbf{-0.2731}} &{\textbf{-0.3255}} \\
\hline
{SAM-ResNet} &{I-FGSM}   &{-0.0164} &{-0.0238} &{-0.0295} \\

{SAM-ResNet} &{MIM}   &{-0.1622} &{-0.1900} &{-0.2258} \\

{SAM-ResNet} &{\textit{SMBEA}}   &{\textbf{-0.2317}} &{\textbf{-0.2996}} &{\textbf{-0.3484}} \\

\toprule[1pt]
\end{tabular}
\end{table}

\begin{figure*}
\centering
\includegraphics[height=0.36\linewidth]{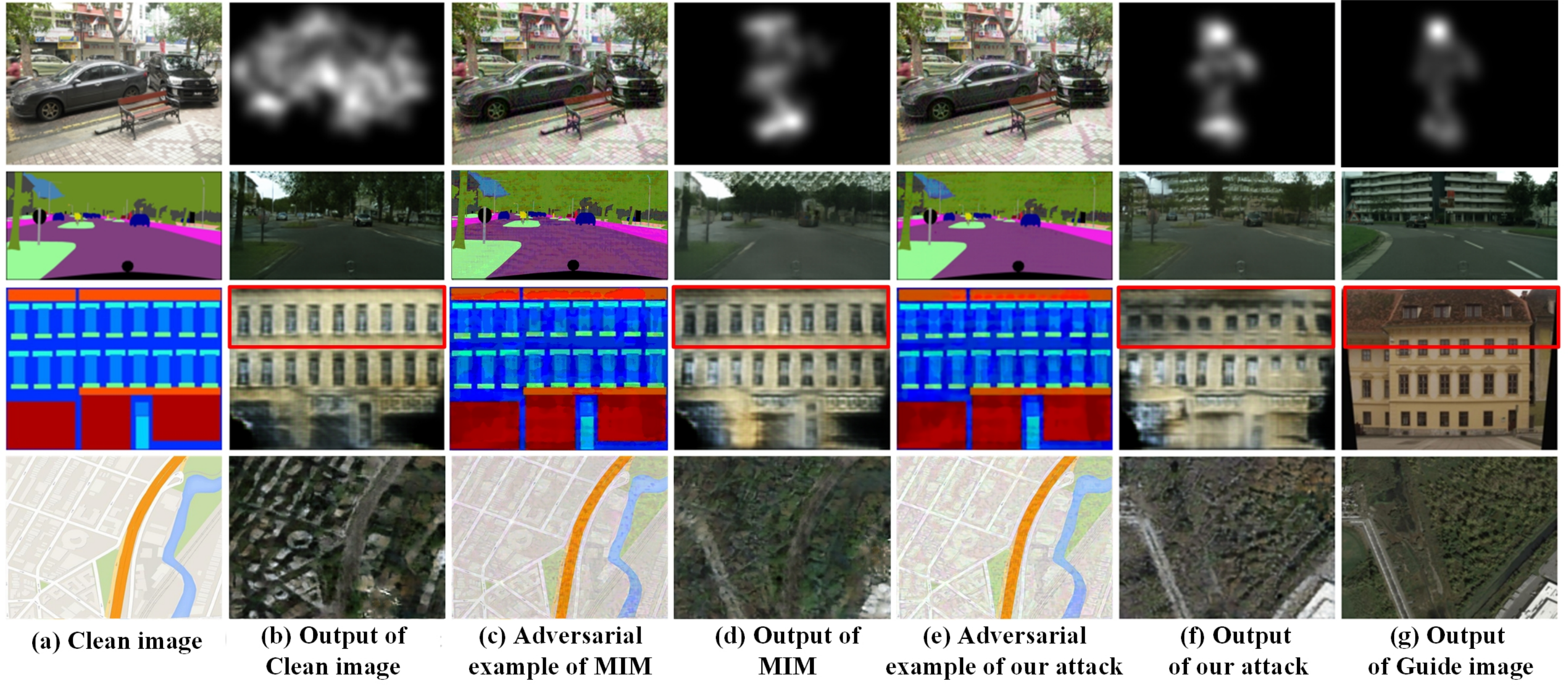}
\vspace{-0.1cm}
\vspace{-0.1cm}
\caption{\footnotesize Qualitative results of \textit{black-box} MIM \cite{MIFGSM} ensemble attack and our attack over multiple pixel-to-pixel vision tasks. }
\label{manytasks}
\end{figure*}

\subsection{Comparison}
In Table.~\ref{Comparison}, we compare our method with other ensemble attacks in the \textit{black-box} setting. These ensemble attacks adopt state-of-the-art gradient
back-propagation attack algorithms. For fair comparison, we use the same perceptual constraint for different competing methods. We can see that our attack achieves the best performance over different datasets.

It was verified that injecting adversarial examples into training set will increase the robustness of deep networks against attacks \cite{fgsm}.
Currently, this adversarial training strategy is the most efficient defense method. In Table.~\ref{AgaintDef}, we compare our method with other attacks against the adversarially trained \textit{black-box} models. Specifically, we keep two adversarially trained models as the hold-out target models, and use the rest source models to craft the adversarial examples. We can see that the adversarially trained models can not defend our attack effectively. 

We further compare our attack with the normal ensemble attack based on PGD \cite{PGD} algorithm against two online \textit{black-box} saliency prediction models, as shown in Fig.~\ref{olattack}. We notice that, by increasing the number of batches, our method misleads the model prediction towards the guide image, while the normal attack fails to fool these models.
Besides, we also compare our method with the \textit{black-box} ensemble attack based on MIM \cite{MIFGSM} algorithm from a qualitative perspective, as shown in Fig.~\ref{manytasks}.
For fair comparison, we apply the same perceptual constraint to different competing methods.
We observe that, the outputs of our attack are more similar to the outputs of the guide images, demonstrating better fooling ability.

\begin{figure}
\centering
\includegraphics[height=0.42\linewidth]{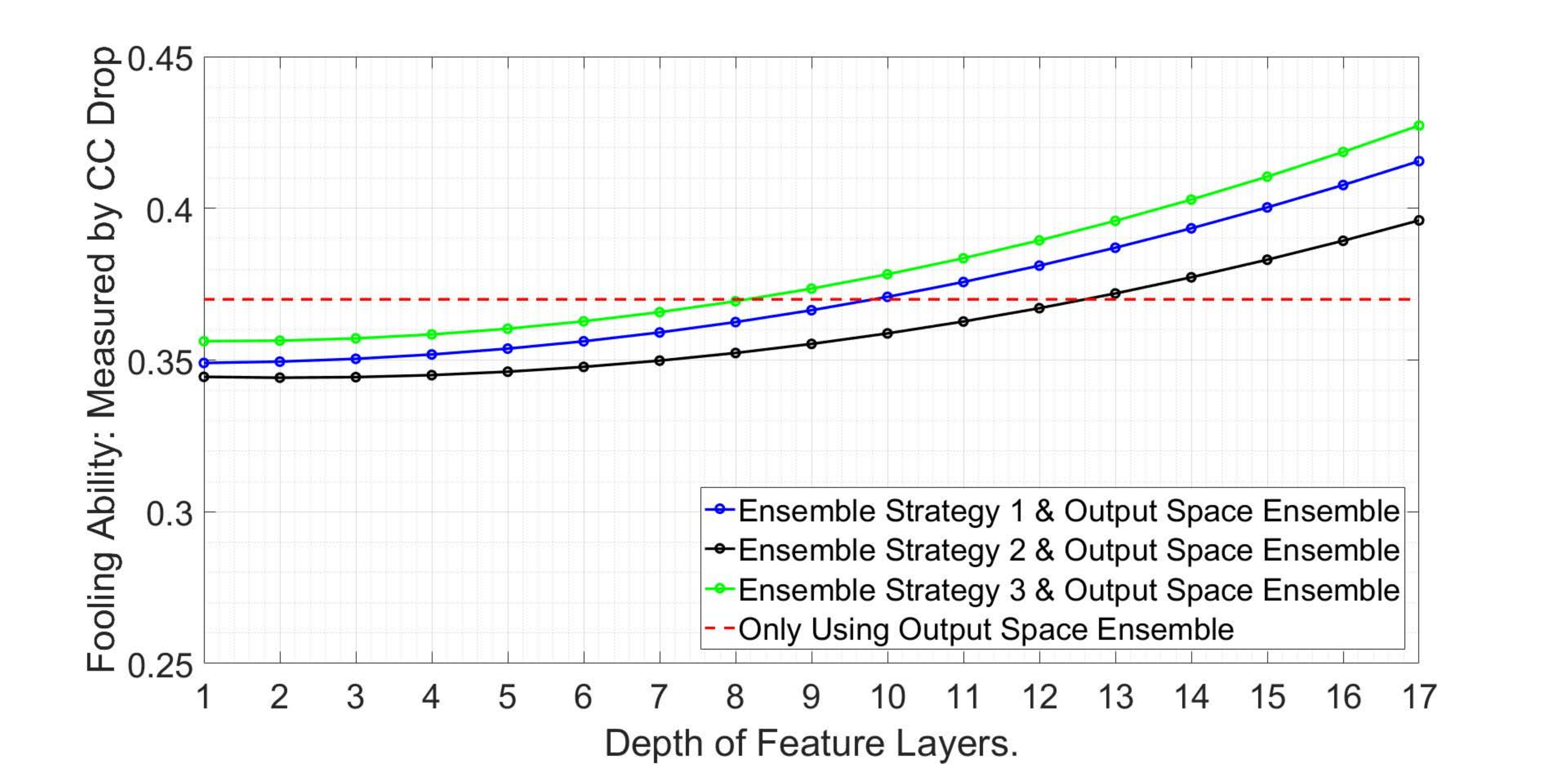}
\caption{\footnotesize The relationship between fooling ability (against the source models) and the depth of feature layers. We compare different ensemble strategies when fusing 4 source models including GazeGAN \cite{gazeganNEW}, Globalpix2pix \cite{pix2pixHD}, SAM-ResNet \cite{SAM}, and SalGAN \cite{SalGAN}. }
\label{featurelayer}
\end{figure}

\begin{figure}
\centering
\includegraphics[height=0.44\linewidth]{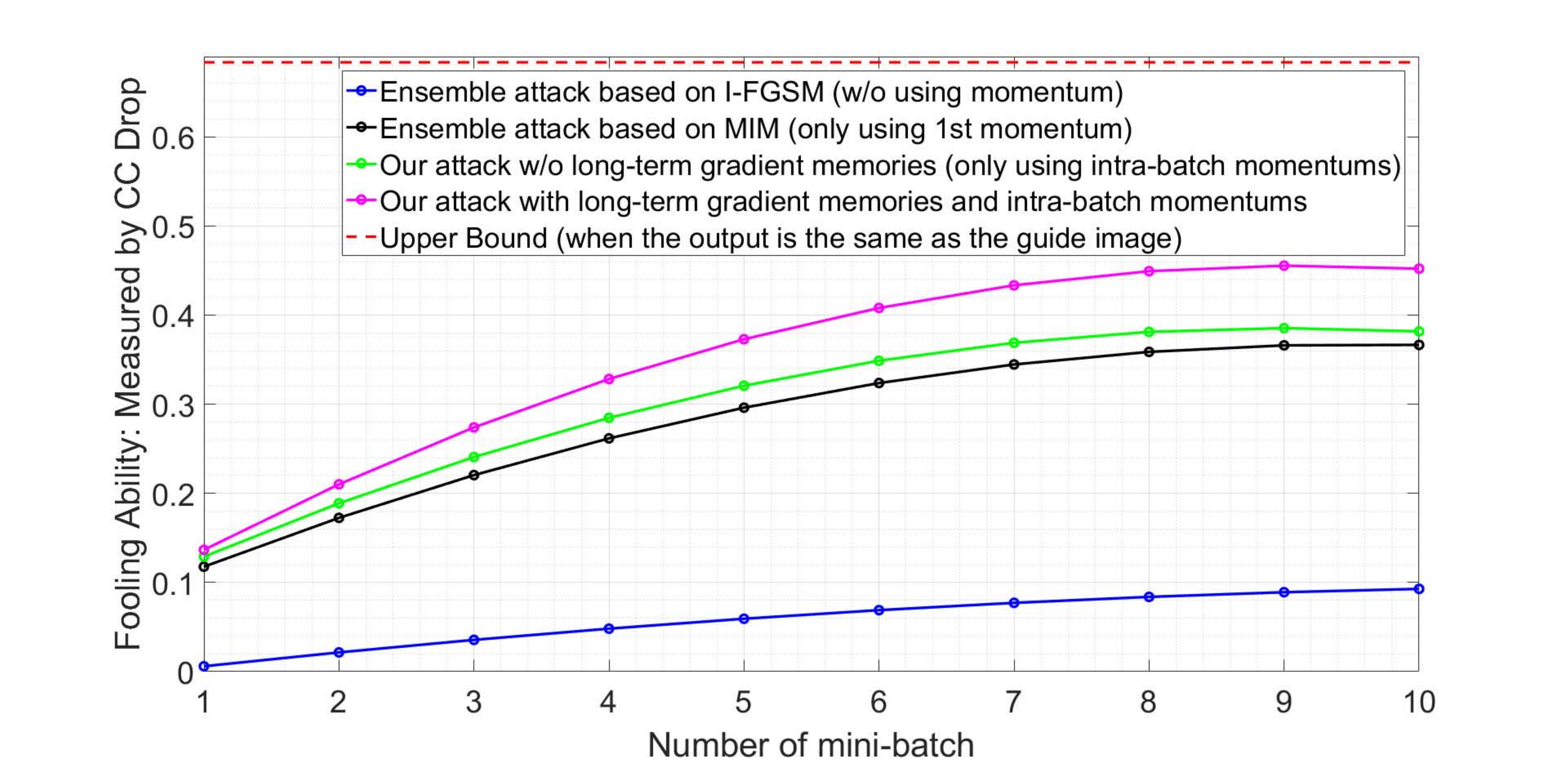}
\caption{\footnotesize Ablation study of our attack. We use the online \textit{black-box} SALICON model as the target model. With the increase of batch number, the perceptual constraint increases (as shown in Step.1 of Algorithm.\ref{alg:inter}). For fair comparison, in the same batch, different competing attacks adopt the same perceptual constraint.}
\label{ablation}
\end{figure}

\subsection{Ablation studies}
In Fig.~\ref{featurelayer}, we explore the relationship between the fooling ability and the depth of feature layers. We introduce 3 new \textit{feature space} ensemble strategies, as shown in Fig.~\ref{ensemblestr}. Here we explain how to select good feature layers to conduct an efficient attack. Experiments indicate that the deeper layers obtain better fooling ability. Besides, the proposed \textit{feature space} ensemble strategies further improve the performance, and the $3_{rd}$ ensemble strategy obtains the best performance.

Next, we explore the contributions of intra-batch momentums and inter-batch ``long-term'' momentums, as shown in Fig.~\ref{ablation}. We notice that, both intra-batch ``short-term'' momentums and inter-batch ``long-term'' gradient memories increase the transferability in the \textit{black-box} setting.

\section{Conclusion}
\label{sec:con}
In this paper, we propose a novel \textit{black-box} attack. Our attack divides a large number of pre-trained source models into several batches. For each batch, we introduce 3 \textit{feature-space} ensemble strategies for improving intra-batch transferability. Besides, we propose a new algorithm that utilizes the ``long-term'' gradient memories. The long-term gradient memories preserve the learned adversarial information and improve inter-batch transferability. Our attack achieves the best performance over multiple pixel-to-pixel datasets, and fools two online \textit{black-box} applications in the real world. We share our code with the community to promote the research on adversarial attack and defense over pixel-to-pixel tasks.

\section{Acknowledgement}
This work was supported in part by the National Science Foundation of China under Grant 61831015, Grant 61771305, and Grant 61927809.

{
\footnotesize
\bibliographystyle{aaai}
\bibliography{ref}

\begin{thebibliography}{}

\bibitem[\protect\citeauthoryear{Alletto \bgroup et al\mbox.\egroup
  }{2016}]{dataattmotodrive}
Alletto, S.; Palazzi, A.; Solera, F.; Calderara, S.; and Cucchiara, R.
\newblock 2016.
\newblock Dr(eye)ve: a dataset for attention-based tasks with applications to
  autonomous and assisted driving.
\newblock In {\em CVPRw}.

\bibitem[\protect\citeauthoryear{Carlini and Wagner}{2017}]{CandW}
Carlini, N., and Wagner, D.
\newblock 2017.
\newblock Towards evaluating the robustness of neural networks.
\newblock In {\em SP}.

\bibitem[\protect\citeauthoryear{Che \bgroup et al\mbox.\egroup
  }{2019}]{gazeganNEW}
Che, Z.; Borji, A.; Zhai, G.; Min, X.; Guo, G.; and Callet, P.
\newblock 2019.
\newblock How is gaze influenced by image transformations? dataset and model.
\newblock {\em TIP}.

\bibitem[\protect\citeauthoryear{Cordts and Omran}{2016}]{Cityspaces}
Cordts, M., and Omran, M.
\newblock 2016.
\newblock The cityscapes dataset for semantic urban scene understanding.
\newblock In {\em CVPR}.

\bibitem[\protect\citeauthoryear{Cornia \bgroup et al\mbox.\egroup
  }{2018}]{SAM}
Cornia, M.; Baraldi, L.; Serra, G.; and et~al.
\newblock 2018.
\newblock Predicting human eye fixations via an lstm-based saliency attentive
  model.
\newblock {\em TIP}.

\bibitem[\protect\citeauthoryear{Dai \bgroup et al\mbox.\egroup }{2017}]{DCN1}
Dai, J.; Qi, H.; Xiong, Y.; Li, Y.; Zhang, G.; Hu, H.; and Wei, Y.
\newblock 2017.
\newblock Deformable convolutional networks.
\newblock In {\em ICCV}.

\bibitem[\protect\citeauthoryear{Dong \bgroup et al\mbox.\egroup
  }{2018}]{MIFGSM}
Dong, Y.; Liao, F.; Pang, T.; Su, H.; Zhu, J.; Hu, X.; and Li, J.
\newblock 2018.
\newblock Boosting adversarial attacks with momentum.
\newblock In {\em CVPR}.

\bibitem[\protect\citeauthoryear{Dong \bgroup et al\mbox.\egroup
  }{2019}]{DongFaceAdv}
Dong, Y.; Su, H.; Wu, B.; Li, Z.; Liu, W.; and Zhang, T.
\newblock 2019.
\newblock Efficient decision-based black-box adversarial attacks on face
  recognition.
\newblock In {\em CVPR}.

\bibitem[\protect\citeauthoryear{Dosovitskiy and Brox}{2016}]{PercepLoss}
Dosovitskiy, A., and Brox, T.
\newblock 2016.
\newblock Generating images with perceptual similarity metrics based on deep
  networks.
\newblock In {\em NeurIPS}.

\bibitem[\protect\citeauthoryear{Duchi, Hazan, and Singer}{2011}]{Adagrad}
Duchi, J.; Hazan, E.; and Singer, Y.
\newblock 2011.
\newblock Adaptive subgradient methods for online learning and stochastic
  optimization.
\newblock {\em JMLR}.

\bibitem[\protect\citeauthoryear{Goodfellow \bgroup et al\mbox.\egroup
  }{2015}]{fgsm}
Goodfellow, I.; Shlens, J.; Szegedy, C.; and Goodfellow, I.
\newblock 2015.
\newblock Explaining and harnessing adversarial examples.
\newblock In {\em ICLR}.

\bibitem[\protect\citeauthoryear{Huang \bgroup et al\mbox.\egroup
  }{2017}]{salicononline}
Huang, X.; Shen, C.; Boix, X.; and Zhao, Q.
\newblock 2017.
\newblock Online saliency prediction system {SALICON}.
\newblock {\em \url{http://salicon.net/demo/}}.

\bibitem[\protect\citeauthoryear{Jiang \bgroup et al\mbox.\egroup
  }{2015}]{SALICONDB}
Jiang, M.; Huang, S.; Duan, J.; and Zhao, Q.
\newblock 2015.
\newblock Salicon: Saliency in context.
\newblock In {\em CVPR}.

\bibitem[\protect\citeauthoryear{Kingma and Ba}{2015}]{Adam}
Kingma, D.~P., and Ba, J.
\newblock 2015.
\newblock Adam: A method for stochastic optimization.
\newblock In {\em ICLR}.

\bibitem[\protect\citeauthoryear{Kummerer}{2017}]{deepgaze2online}
Kummerer, M.
\newblock 2017.
\newblock Online saliency prediction system {Deepgaze-II}.
\newblock {\em \url{https://deepgaze.bethgelab.org/}}.

\bibitem[\protect\citeauthoryear{Kurakin \bgroup et al\mbox.\egroup
  }{2016}]{itefgsm}
Kurakin, A.; Goodfellow, I.; Bengio, S.; and Bengio, S.
\newblock 2016.
\newblock Adversarial examples in the physical world.
\newblock In {\em ICLRw}.

\bibitem[\protect\citeauthoryear{Liu \bgroup et al\mbox.\egroup
  }{2017}]{EnsemAttack}
Liu, Y.; Chen, X.; Liu, C.; and Song, D.
\newblock 2017.
\newblock Delving into transferable adversarial examples and black-box attacks.
\newblock In {\em ICLR}.

\bibitem[\protect\citeauthoryear{Madry \bgroup et al\mbox.\egroup }{2018}]{PGD}
Madry, A.; Makelov, A.; Schmidt, L.; Tsipras, D.; and Vladu, A.
\newblock 2018.
\newblock Towards deep learning models resistant to adversarial attacks.
\newblock In {\em ICLR}.

\bibitem[\protect\citeauthoryear{Mopuri, Ganeshan, and
  Radhakrishnan}{2018}]{pamifeature}
Mopuri, K.~R.; Ganeshan, A.; and Radhakrishnan, V.~B.
\newblock 2018.
\newblock Generalizable data-free objective for crafting universal adversarial
  perturbations.
\newblock {\em TPAMI}.

\bibitem[\protect\citeauthoryear{Pan \bgroup et al\mbox.\egroup
  }{2017}]{SalGAN}
Pan, J.; Canton, C.; McGuinness, K.; Connor, N.; Torres, J.; Sayrol, E.; and
  Nieto, X.
\newblock 2017.
\newblock Salgan: Visual saliency prediction with generative adversarial
  networks.
\newblock In {\em CoRR:1701.01081}.

\bibitem[\protect\citeauthoryear{Papernot \bgroup et al\mbox.\egroup
  }{2016a}]{CleverHans}
Papernot, N.; Goodfollow, I.; Sheatsley, R.; Feinman, R.; and McDaniel, P.
\newblock 2016a.
\newblock cleverhans v2. 0.0: an adversarial machine learning library.
\newblock In {\em arXiv preprint}.

\bibitem[\protect\citeauthoryear{Papernot \bgroup et al\mbox.\egroup
  }{2016b}]{DefeseDis}
Papernot, N.; McDaniel, P.; Wu, X.; Jha, S.; and Swami, A.
\newblock 2016b.
\newblock Distillation as a defense to adversarial perturbations against deep
  neural networks.
\newblock In {\em SP}.

\bibitem[\protect\citeauthoryear{Papernot}{2017}]{PapernotBB}
Papernot, N.
\newblock 2017.
\newblock Practical black-box attacks against machine learning.
\newblock In {\em ACM ACCCS}.

\bibitem[\protect\citeauthoryear{Qian}{1999}]{MSGD}
Qian, N.
\newblock 1999.
\newblock On the momentum term in gradient descent learning algorithms.
\newblock {\em Neural networks}.

\bibitem[\protect\citeauthoryear{Sharif \bgroup et al\mbox.\egroup
  }{2016}]{AdvGlass}
Sharif, M.; Bhagavatula, S.; Bauer, L.; and Reiter, M.~K.
\newblock 2016.
\newblock Accessorize to a crime: Real and stealthy attacks on state-of-the-art
  face recognition.
\newblock In {\em ACM SIGSAC}.

\bibitem[\protect\citeauthoryear{Szegedy \bgroup et al\mbox.\egroup
  }{2014}]{bfgs}
Szegedy, C.; Zaremba, W.; Sutskever, I.; Bruna, J.; Erhan, D.; Goodfellow,
  I.~J.; and Fergus, R.
\newblock 2014.
\newblock Intriguing properties of neural networks.
\newblock In {\em ICLR}.

\bibitem[\protect\citeauthoryear{Tieleman and Hinton}{2012}]{RMSProp}
Tieleman, T., and Hinton, G.
\newblock 2012.
\newblock Lecture 6.5-rmsprop: Divide the gradient by a running average of its
  recent magnitude.
\newblock {\em COURSERA: Neural networks for machine learning}.

\bibitem[\protect\citeauthoryear{Tylecek}{2013}]{Facades}
Tylecek, R.
\newblock 2013.
\newblock Spatial pattern templates for recognition of objects with regular
  structure.
\newblock In {\em GCPR}.

\bibitem[\protect\citeauthoryear{Wang \bgroup et al\mbox.\egroup
  }{2018}]{pix2pixHD}
Wang, T.~C.; Liu, M.~Y.; Zhu, J.~Y.; Tao, A.; Kautz, J.; and Catanzaro, B.
\newblock 2018.
\newblock High-resolution image synthesis and semantic manipulation with
  conditional gans.
\newblock In {\em CVPR}.

\bibitem[\protect\citeauthoryear{Wei \bgroup et al\mbox.\egroup
  }{2019}]{GANimg}
Wei, X.; Liang, S.; Chen, N.; and Cao, X.
\newblock 2019.
\newblock Transferable adversarial attacks for image and video object
  detection.
\newblock In {\em IJCAI}.

\bibitem[\protect\citeauthoryear{Xie \bgroup et al\mbox.\egroup
  }{2017}]{iccvsegadv}
Xie, C.; Wang, J.; Zhang, Z.; Zhou, Y.; and an~A.~Yuille, L.~X.
\newblock 2017.
\newblock Adversarial examples for semantic segmentation and object detection.
\newblock In {\em ICCV}.

\bibitem[\protect\citeauthoryear{Yang and Hsu}{2017}]{smautodrive}
Yang, S., and Hsu, Y.
\newblock 2017.
\newblock Full speed region sensorless drive of permanent-magnet machine
  combining saliency-based and back-emf-based drive.
\newblock {\em TIE}.

\bibitem[\protect\citeauthoryear{Yu and Koltun}{2016}]{DilatedConv}
Yu, F., and Koltun, V.
\newblock 2016.
\newblock Multi-scale context aggregation by dilated convolutions.
\newblock In {\em ICLR}.

\bibitem[\protect\citeauthoryear{Zhao, Dheeru, and Sameer}{2018}]{LatentGANAdv}
Zhao, Z.; Dheeru, D.; and Sameer, S.
\newblock 2018.
\newblock Generating natural adversarial examples.
\newblock In {\em ICLR}.

\end{thebibliography}
}
\end{document}